\documentclass{article}
\usepackage[T1]{fontenc}
\usepackage{iftex}
\ifXeTeX
  \usepackage{newunicodechar}
  \newunicodechar{–}{\textendash}
\fi
\usepackage{iclr2027_conference,times}
\iclrfinalcopy
\usepackage{amsmath,amssymb,amsthm}
\usepackage{graphicx}
\usepackage{xcolor,colortbl}
\definecolor{tableblue}{RGB}{232,243,252}
\usepackage{booktabs}
\usepackage{multirow}
\usepackage{placeins}
\usepackage{algorithm}
\usepackage{algpseudocode}
\usepackage{hyperref}
\usepackage{url}
\usepackage{wrapfig}
\usepackage{enumitem}

\newcommand{\dloop}{D-Loop}
\newcommand{\mask}{\textsc{[Mask]}}
\newcommand{\KL}{D_{\mathrm{KL}}}
\newcommand{\E}{\mathbb{E}}
\newcommand{\Ind}{\mathbf{1}}
\definecolor{citeblue}{RGB}{100,160,210}

\title{\raggedright D-Loop: Looped Diffusion Drafting for \mbox{Speculative} Decoding}
\author{\parbox{\dimexpr\textwidth-2\tabcolsep\relax}{\normalfont\centering
\mbox{Kecheng Chen\textsuperscript{1}}, \mbox{Yuyang He\textsuperscript{2}}, \mbox{Cheng Gong\textsuperscript{1}}, \mbox{Hui Liu\textsuperscript{1}},
\mbox{Guoping Long\textsuperscript{3}}, \mbox{Jiajun Li\textsuperscript{3}},\\
\mbox{Shi Wu\textsuperscript{3}}, \mbox{Suiyun Zhang\textsuperscript{3}},
\mbox{Haoliang Li\textsuperscript{1}}, \mbox{Ziru Liu\textsuperscript{3,*}}, \mbox{Rui Liu\textsuperscript{3,*}}\\[0.5em]
\textsuperscript{1}City University of Hong Kong\\
\textsuperscript{2}The Chinese University of Hong Kong\\
\textsuperscript{3}Huawei Research\\[0.3em]
{\small\textsuperscript{*}Corresponding authors.}
}}
\hypersetup{ colorlinks=true,
    citecolor=citeblue,
    linkcolor=black,
    urlcolor=black,pdfauthor={Kecheng Chen, Yuyang He, Cheng Gong, Hui Liu, Guoping Long, Jiajun Li, Shi Wu, Suiyun Zhang, Haoliang Li, Ziru Liu, Rui Liu},pdftitle={D-Loop: Looped Diffusion Drafting for Speculative Decoding}}

\begin{document}
\maketitle
\fancyhead{}
\renewcommand{\headrulewidth}{0pt}
\begin{abstract}
Block diffusion accelerates speculative decoding by drafting multiple
tokens in one forward pass. However, each position predicts a marginal
distribution without observing earlier proposed tokens, limiting draft
quality and acceptance length. We identify a concrete failure, the
\emph{repetition trap}, in which neighboring positions produce redundant
copies of the same token. We explain this tendency theoretically and
empirically examine its association with shorter accepted drafts.
Recent methods refine marginal predictions with an additional causal
head or a separately trained drafter, increasing parameter storage and
introducing separate training objectives. We instead propose \dloop{},
which introduces \emph{intra-block causal conditioning} within the
original diffusion drafter without additional model components. Inspired
by semi-autoregressive generation and parameter sharing, \dloop{} reuses
the same backbone across looped passes. The first pass proposes a block,
and the second conditions on a selected prefix to regenerate the suffix
in parallel. A complementary prefix--suffix objective trains the shared
drafter for both anchor-only prefix prediction and prefix-conditioned
suffix prediction. Across eight math, code, and chat benchmarks,
\dloop{} can beat DFlash and DSpark on Qwen3-4B  and Qwen3-8B with obvious gains.
\end{abstract}

\section{Introduction}
\label{sec:introduction}

Large language models generate outputs through a sequence of 
autoregressive forward passes. Speculative decoding reduces this cost by
using a lightweight draft model (\textit{a.k.a.} drafter) to propose several tokens and a stronger target
model to verify them in parallel
\citep{leviathan2023speculative,chen2023sampling}. Its effectiveness depends
on both the cost of producing a proposal and the length of the prefix that
survives verification. Compared to a small autoregressive drafter, diffusion drafting offers an appealing way to reduce proposal latency. As a pioneer,
DFlash \citep{chen2026dflash}, inspired by diffusion language models (DLMs), conditions a lightweight block diffusion model
on target-model features and predicts a block of masked tokens in one
forward pass. This parallel computation permits a deeper drafter than is
typically affordable in token-by-token speculation. However, its central modeling
limitation is the lack of \emph{intra-block causal conditioning}, \textit{i.e.,}
each position predicts a marginal distribution without observing the tokens
realized earlier in the same block, causing compromised proposal quality and limiting the accepted draft length.


In this paper, we identify a concrete failure of marginal modeling that we call the
\emph{repetition trap}, where neighboring positions produce redundant
copies of the same token within a draft block. In
Section~\ref{sec:observations}, we provide a theoretical explanation of
why marginal predictions can overproduce such repetitions and quantify
their prevalence in actual draft outputs. These findings highlight the need to improve
marginal draft predictions by accounting for dependencies among proposed
tokens.

To this end, several recent methods improve marginal draft predictions by introducing
additional refinement modules. For example, DSpark \citep{cheng2026dspark} augments a
parallel drafter with lightweight causal heads or refiners, while D$^2$SD
\citep{zhang2026d2sd} introduces a separately trained diffusion drafter
for prefix-conditioned redrafting. These approaches improve draft quality
but require additional model parameters. Frameworks such as DSpark also
combine multiple training objectives. This raises a simple question:
\emph{can we alleviate the limitations of marginal modeling within the
original drafter, without introducing any additional components?}
Such a design would be attractive for memory-constrained deployment by
avoiding the storage of a separate refinement module or a second set of
drafter weights. It also motivates a simpler training formulation that
supports refinement without auxiliary head-specific objectives.

Notably, an well-established approach to strengthening generative quality of DLMs is semi-autoregressive generation, where each block conditions on previously generated blocks while retaining parallel prediction within a block \citep{arriola2025block}. Such an inter-block causal conditioning is devised without adding a
separate model component. Drawing on this principle and parameter sharing
in recurrent-depth models \citep{geiping2025recurrent}, we propose
\dloop{}, a looped diffusion drafter that operates on discrete proposal
states. The core idea of \dloop{} is to introduce \emph{intra-block causal
conditioning} through looped shared-parameter forward passes. An earlier
pass proposes a token block, and a later pass conditions on a selected
prefix to regenerate the suffix in parallel. Both passes reuse the full
drafter with the same parameters. 

Our preliminary experiments found that the vanilla full-masking loss adopted by DFlash and DSpark is insufficient to enable looped forward passes. To make one drafter effective at both initial proposal and prefix-conditioned refinement, we develop a
\emph{complementary prefix--suffix training objective}. An anchor-only
view supervises the prefix, and a revealed-prefix view supervises the
suffix. Both views update shared parameters under one normalized objective. Despite an additional looped forward pass with a double drafting time (see Table~\ref{tab:latency}), \dloop{} still achieves better speedup performance (see  Tables~\ref{tab:main} and~\ref{tab:qwen3-8b}) due to its increased acceptance length compared to DFlash and DSpark.

Our contributions are threefold:
\begin{itemize}[leftmargin=*]
    \item We identify a \emph{repetition trap} in block diffusion drafting
    for speculative decoding and provide a theoretical explanation of how
    marginal modeling can produce it. We quantify repetition in actual draft
    outputs and examine its association with shorter accepted drafts.
    \item Without adding a causal head or a second drafter, we propose \dloop{},
    which introduces \emph{intra-block causal conditioning} within the
    diffusion drafter itself through shared-parameter looped passes.
    We train both passes with a \emph{complementary prefix--suffix objective}.
    \item Experiments on math, code, and chat tasks with Qwen3-4B and
    Qwen3-8B demonstrate \dloop{}'s effectiveness across model scales and
    task domains.
\end{itemize}

\section{Theoretical Analyses and Empirical Observations}
\label{sec:observations}
\label{sec:repetition}

By examining actual token proposals from marginal block diffusion drafting,
we identify a systematic failure: the \emph{repetition trap}, in which
multiple positions emit the same token consecutively. We characterize this
phenomenon through empirical observations and theoretical analysis. 

\paragraph{Notation.}
Let $C$ denote the verified context together with its available target
features and a known anchor token $x_0$. A draft block contains $B=m+1$
positions: the anchor and $m$ proposed tokens $X_{1:m}$. Let
$p(x_{1:m}\mid C)$ be the target continuation distribution.

\paragraph{Marginal block drafting.}

Consider a block diffusion drafter that predicts all masked proposal
positions in parallel from an anchor-only state. With independent token
sampling, its output distribution factorizes as
\begin{equation}
    q^{(0)}_\theta(x_{1:m}\mid C)
    =\prod_{k=1}^{m}q_{\theta,k}(x_k\mid C,z^{(0)}),
    \qquad z^{(0)}=(x_0,\mask,\ldots,\mask).
    \label{eq:parallel}
\end{equation}
DFlash \citep{chen2026dflash} instantiates this factorization by generating
the entire candidate block in a single drafter forward pass before target
verification. We refer to this setting as \emph{one-pass block diffusion
drafting}: one denoising pass fills all masked proposal positions, with no
further refinement of the proposed tokens. DSpark \citep{cheng2026dspark}
also retains a single parallel backbone pass, but adds a sequential head
whose token conditioning goes beyond Equation~\ref{eq:parallel}.

In Equation~\ref{eq:parallel}, each factor predicts a positional marginal
without observing the tokens realized at other positions, even when the logits share attention and
hidden representations. Prior work discusses this missing dependence and
its consequences for draft acceptance.
However, consecutive repetition within raw draft blocks has not been
specifically characterized in these studies.

\paragraph{An observed failure: the repetition trap.}
We observe that multiple positions in a single draft block can emit the
same token consecutively, forming long repeated runs. We call this behavior
the \emph{repetition trap}. Table~\ref{tab:repetition} quantifies this
pattern in raw DFlash proposals. MT-Bench exhibits a higher adjacent-repeat
rate than GSM8K ($42.6\%$ versus $17.4\%$) and longer repeated runs
($4.7$ versus $2.5$ tokens), with LiveCodeBench between them on both metrics.
Consecutive repeated content words occur in $72.5\%$--$94.5\%$ of blocks,
showing that the phenomenon extends beyond whitespace repetition.
The examples illustrate two forms: \emph{position confusion} duplicates
a word before continuing, while \emph{endless} repetition persists
through the shown suffix.
This observation identifies
a concrete failure associated with missing output dependencies. To our knowledge,
this is the first report specifically documenting this phenomenon in block
diffusion drafting for speculative decoding. Note that repetition has also
been observed in non-autoregressive neural machine translation
\citep{gu2018nonautoregressive,ran2020recover}. We analyze its origin in
marginal drafting and its cost under prefix verification.

\begin{table}[t]
\centering
\caption{DFlash repetition statistics
and illustrative examples. Acc.\ Len.\ uses Qwen3-8B at $T=1$
(Table~\ref{tab:qwen3-8b}). LCB denotes LiveCodeBench.}
\label{tab:repetition}
\begin{minipage}[t]{0.52\textwidth}
\vspace{0pt}
\centering
\fontsize{8}{9.5}\selectfont
\setlength{\tabcolsep}{2.5pt}
\renewcommand{\arraystretch}{1.4}
\begin{tabular}{@{}lcccc@{}}
\toprule
Dataset
& \shortstack{Adjacent\\repeat}
& \shortstack{Mean longest\\run}
& \shortstack{Content-repeat\\blocks}
& \shortstack{Acc.\ Len.} \\
\midrule
\rowcolor{tableblue}
GSM8K & $17.4\%$ & 2.5 & $72.5\%$ & 5.28 \\
LCB & $29.4\%$ & 3.6 & $84.9\%$ & 3.26 \\
\rowcolor{tableblue}
MT-Bench & $42.6\%$ & 4.7 & $94.5\%$ & 3.10 \\
\bottomrule
\end{tabular}
\end{minipage}\hfill
\begin{minipage}[t]{0.45\textwidth}
\vspace{0pt}
\setlength{\fboxsep}{5pt}
\fcolorbox{black}{black!6}{\begin{minipage}{\dimexpr\linewidth-2\fboxsep-2\fboxrule\relax}
\fontsize{8}{10}\selectfont
\raggedright
\textbf{Type 1: Position confusion}\par\smallskip
\begin{tabular}{@{}l@{\hspace{3pt}}l@{}}
Gold: & \texttt{how many laps Beth} \\
Draft: & \texttt{how many \textbf{laps laps} Beth}
\end{tabular}
\par\medskip\hrule\medskip
\textbf{Type 2: Endless}\par\smallskip
Draft: \texttt{subtracting \textbf{emails emails} \ldots}
\end{minipage}}
\end{minipage}
\end{table}

\paragraph{Why marginal predictions can overproduce repetitions.}
For a fixed context $c$, write $p_c=p(X_{1:m}\mid c)$. Even exact
marginals leave a product-approximation gap
\citep{huang2022learning}:
\begin{equation}
    \min_{q=\prod_k q_k}\KL(p_c\|q)
    =\sum_k H(X_k\mid c)-H(X_{1:m}\mid c)
    =\sum_k I(X_k;X_{<k}\mid c).
    \label{eq:tc}
\end{equation}
The optimum is $q^\star=\prod_k p(X_k\mid c)$: each cross-entropy term is
minimized by its target marginal, and the entropy chain rule gives the
last equality. Each mutual-information term measures how much observing
the realized prefix reduces uncertainty about the next token. Marginal
drafting discards precisely this information, including \textit{whether a word
has already been emitted, leading to a potential repetition risk.}
To make this comparison explicit, let $\pi_k(v)=p(X_k=v\mid c)>0$. We compare the probability of
the adjacent repetition $X_k=X_{k+1}=v$ under the independent-marginal
distribution $q^\star$ with that under the target joint distribution $p$.
Their difference is
\begin{equation}
\begin{aligned}
    \Delta_{k,v}
    &=q^\star(X_k=X_{k+1}=v\mid c)-p(X_k=X_{k+1}=v\mid c)\\
    &=\underbrace{\pi_k(v)\pi_{k+1}(v)}_{\text{independent marginals}}
      -\underbrace{\pi_k(v)p(X_{k+1}=v\mid X_k=v,c)}_{\text{target joint distribution}}\\
    &=\pi_k(v)\big[\pi_{k+1}(v)-p(X_{k+1}=v\mid X_k=v,c)\big].
\end{aligned}
    \label{eq:repeat-excess}
\end{equation}
The first product uses the next position's marginal probability, without
observing whether the preceding position selected $v$. The second product
is the target joint probability written using the chain rule: it conditions
the next position on $X_k=v$, retaining the target's adjacent-token
dependence. As we can see, whenever observing $v$ lowers the probability of
immediately emitting it again below $\pi_{k+1}(v)$, we have
$\Delta_{k,v}>0$. This reveals that independent marginal predictions assign a higher
probability to the adjacent repetition ($vv$) at positions $k$ and $k+1$
than the target joint distribution does, even when the marginals are exact. Thus,
uncertainty about where a word belongs can become redundant copies when
positions cannot condition on realized tokens.

\paragraph{Implications for causal drafting.}
Speculative verification accepts only a contiguous prefix. The repetition
trap can therefore compound suffix decay in accepted drafts: an erroneous
repeat can trigger early rejection and discard every subsequent proposal.


%

\section{D-Loop: Looped Diffusion Drafting}
\label{sec:method}

\begin{figure}[t]
    \centering
    \includegraphics[width=\linewidth]{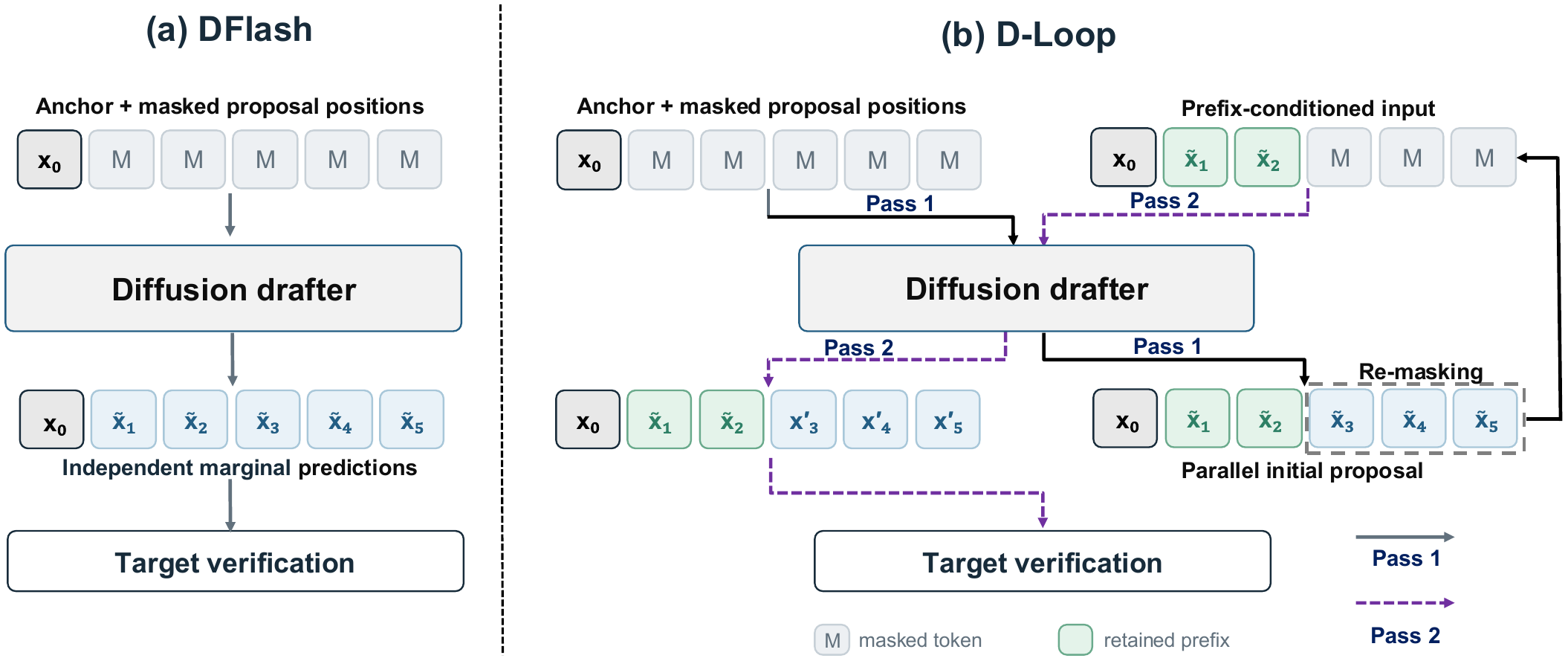}
    \caption{\textbf{Overview of DFlash and \dloop{}.}
    (a) DFlash predicts all proposal tokens in one parallel pass.
    (b) \dloop{} retains a selected prefix from the first pass, remasks the
    suffix, and invokes the same drafter to regenerate the suffix conditioned
    on the retained tokens. Both passes share all draft parameters and use
    target features from the verified context. Target verification follows
    the second pass. One selected prefix is shown for clarity.}
    \label{fig:framework}
\end{figure}

\textbf{Overall.} Instead of using any extra causal head, we propose \dloop{}, which leverages intra-block causal conditioning within the diffusion
drafter itself. An initial forward pass instantiates a provisional token
prefix. Then, a second pass uses that prefix to regenerate the suffix through
the same backbone. The loop operates on discrete proposal states and reuses
DFlash's target-feature conditioning and token prediction interface.
Figure~\ref{fig:framework} contrasts the two drafting procedures.

\subsection{From One-Pass Marginals to a Prefix-Conditioned Loop}
\label{sec:loop}

Block diffusion \citep{arriola2025block} follows a semi-autoregressive
factorization: it generates blocks sequentially, conditioning each block
on previously generated blocks, while using diffusion to predict tokens
in parallel within a block. Inspired by this block-causal structure, we
introduce a similar dependence \emph{inside a single speculative block}.
An earlier pass proposes a prefix, and a later pass conditions on that
prefix to regenerate the suffix. This yields \emph{intra-block causal
conditioning} while retaining parallel prediction within each stage.

Starting from $z^{(0)}$ in Equation~\ref{eq:parallel}, the two passes
reuse the same drafter:
\begin{equation}
    z^{(0)}\xrightarrow{q_\theta}\widetilde{x}_{1:m},
    \qquad
    \underbrace{(x_0,\widetilde{x}_{1:\ell},\mask,\ldots,\mask)}_
    {z^{(\ell)}(\widetilde{x}_{1:\ell})}
    \xrightarrow{q_\theta}y_{\ell+1:m}.
    \label{eq:canvas}
\end{equation}
The first pass produces all $m$ proposal tokens $\widetilde{x}_{1:m}$
in parallel. We select a length $\ell\in\{0,\ldots,m-1\}$ and retain
the first $\ell$ tokens as $\widetilde{x}_{1:\ell}$. The second input
$z^{(\ell)}(\widetilde{x}_{1:\ell})$ contains the known anchor $x_0$,
this visible prefix, and $m-\ell$ mask tokens. The second pass predicts
a new suffix $y_{\ell+1:m}$ in parallel, yielding the final proposal
$y=(\widetilde{x}_{1:\ell},y_{\ell+1:m})$. Both arrows denote a drafter
forward pass followed by token selection, conditioned on the same
verified context $C$. The superscript $(\ell)$ counts visible proposal
tokens, excluding the anchor.

\paragraph{Selecting candidate prefixes.}
To select where to truncate $\widetilde{x}_{1:m}$, we adopt the
confidence-guided top-$K$ prefix selection 
\citep{zhang2026d2sd}. Let
$c_k=q_{\theta,k}(\widetilde{x}_k\mid C,z^{(0)})$ be the confidence
assigned to the proposed token at position $k$. Each prefix length is
scored by $
    r_\ell=\left(\prod_{k=1}^{\ell}c_k\right)(1-c_{\ell+1}),
    \qquad \ell\in\{0,\ldots,m-1\}.$
This score favors a confident prefix followed by a low-confidence token,
identifying a candidate boundary for suffix regeneration. We select the
top-$K$ lengths by $r_\ell$. For each selected $\ell$, we retain exactly
the first $\ell$ tokens of the original proposal,
$\widetilde{x}_{1:\ell}=(\widetilde{x}_1,\ldots,\widetilde{x}_\ell)$,
and mask positions $\ell+1$ through $m$. The second pass then regenerates
the suffix from $z^{(\ell)}(\widetilde{x}_{1:\ell})$.

\begingroup
\setlength{\intextsep}{3pt}
\paragraph{Alleviating the repetition trap.}
With $K=1$, the second pass reduces adjacent-repeat rates by $73.7\%$,
$83.0\%$, and $79.8\%$ on GSM8K, LCB, and MT-Bench
(Table~\ref{tab:loop-repetition}). In the observed runs, repeated tokens
typically have lower confidence than the preceding token. Hence,
\begin{wraptable}{r}{0.32\textwidth}
\centering
\small
\setlength{\abovecaptionskip}{0pt}
\caption{Adjacent-repeat rates.}
\label{tab:loop-repetition}
\setlength{\tabcolsep}{5pt}
\renewcommand{\arraystretch}{1.15}
\begin{tabular}{@{}lcc@{}}
\toprule
Dataset & Pass 1 & Pass 2 \\
\midrule
GSM8K & $19.4\%$ & $\mathbf{5.1\%}$ \\
LCB & $28.8\%$ & $\mathbf{4.9\%}$ \\
MT-Bench & $44.5\%$ & $\mathbf{9.0\%}$ \\
\bottomrule
\end{tabular}
\end{wraptable}
repetition tends to leave them in the suffix for
regeneration. The visible prefix lets the drafter account for tokens
already emitted, reducing redundant copies. Correcting such repeats
can avoid early rejection and preserve acceptance of later tokens.
These results show that the loop mitigates the failure
in Section~\ref{sec:repetition}, supporting repetition reduction as one
mechanism behind longer accepted drafts. 
\par\endgroup

For a fixed split selected independently of proposal tokens, the induced
two-stage proposal distribution has the form
\begin{equation}
    q_\theta^{(\ell)}(x_{1:m}\mid C)
    =\underbrace{\prod_{k=1}^{\ell}q_{\theta,k}(x_k\mid C,z^{(0)})}_{\text{parallel prefix}}
    \underbrace{\prod_{k=\ell+1}^{m}
    q_{\theta,k}(x_k\mid C,z^{(\ell)}(x_{1:\ell}))}_{\text{prefix-conditioned parallel suffix}}.
    \label{eq:loop-factor}
\end{equation}
The first product is the probability of the retained prefix $x_{1:\ell}$
under the first-pass predictions. The second is the probability of the
regenerated suffix $x_{\ell+1:m}$ given that same prefix as visible input.
Thus, the final proposal combines the first-pass prefix with the
second-pass suffix, replacing the original suffix. Each stage predicts
its tokens in parallel, but the suffix now conditions on the prefix.

\paragraph{Intra-block causal conditioning.}
Equation~\ref{eq:loop-factor} creates the directed dependence
$X_{1:\ell}\rightarrow X_{\ell+1:m}$ through revealed tokens. It remains
parallel within the prefix and within the regenerated suffix. The causal
dependence comes from making proposed prefix tokens visible to the next
pass through hard-token re-embedding. DFlash's bidirectional attention is
retained over the visible prefix and masked suffix. Meanwhile, no additional
token-level causal attention mask is required. Both stages belong to the
same draft block, whose tokens remain provisional until target verification.

\paragraph{Parameter reuse versus a new sequential head.}
DSpark supplies realized-token dependencies through a separate Markov or
RNN head \citep{cheng2026dspark}. \dloop{} instead invokes its existing
backbone again on a different visible state. This shares the motivation
of recurrent-depth computation \citep{geiping2025recurrent}, but does not
recur over a continuous hidden state or assume arbitrary-depth
extrapolation. Both passes can use one in-memory copy of the draft weights.
A theoretical comparison with one-pass marginal drafting is provided in
Appendix~\ref{sec:information}.

\subsection{Complementary Prefix--Suffix Training}
\label{sec:training}

Let $x_{1:m}$ be a training continuation generated by the target model.
For each training block, we draw a prefix length $\ell\sim\pi_\beta$
to determine how many ground-truth proposal tokens to reveal in the
suffix view. The sampling probability is
\begin{equation}
    \pi_\beta(\ell)=\frac{(1-\beta)\beta^\ell}{1-\beta^m},
    \qquad \ell\in\{0,\ldots,m-1\},\quad 0<\beta<1.
    \label{eq:prior}
\end{equation}
Here $\pi_\beta(\ell)$ is the probability of selecting length $\ell$,
and $\beta$ controls how quickly this probability decreases with length.
Smaller $\beta$ favors shorter prefixes more strongly. As $\beta$
approaches one, the distribution becomes nearly uniform over the allowed
lengths. Sampling from this distribution trains the drafter under varied
amounts of visible prefix information. It is a training distribution,
not an estimate of target acceptance or the inference-time selection rule.

Each sampled block yields two views. The \emph{suffix view} exposes the
ground-truth anchor and prefix, supervising only the masked suffix. The
\emph{prefix view} exposes only the anchor, supervising the complementary
prefix positions. Write
\begin{equation}
\begin{aligned}
    h^S_{ik}&=-\log q_{\theta,k}(x_{ik}\mid C_i,z_i^{(\ell_i)}(x_{i,1:\ell_i})),
    &\quad h^P_{ik}&=-\log q_{\theta,k}(x_{ik}\mid C_i,z_i^{(0)}),\\
    w^S_{ik}&=v_{ik}\Ind\{k>\ell_i\}e^{-(k-\ell_i-1)/\tau_s},
    &\quad w^P_{ik}&=v_{ik}\Ind\{1\leq k\leq\ell_i\}e^{-(k-1)/\gamma},
\end{aligned}
\label{eq:weights}
\end{equation}
where $i$ indexes sampled blocks and $v_{ik}$ excludes invalid or padded
positions. The positive scales $\tau_s$ and $\gamma$ control loss decay
within the suffix and prefix views, respectively. The anchor is never
supervised. We optimize the minibatch
objective
\begin{equation}
    \mathcal{L}_{\mathrm{comp}}(\theta)
    =\frac{\sum_i\sum_{k=1}^{m}
    \left(w^P_{ik}h^P_{ik}+w^S_{ik}h^S_{ik}\right)}
    {\sum_i\sum_{k=1}^{m}(w^P_{ik}+w^S_{ik})+\epsilon}.
    \label{eq:loss}
\end{equation}
Here $\epsilon>0$ is a small numerical constant that prevents division
by zero when the total valid weight vanishes. This is one weighted
cross-entropy objective over complementary
prediction conditions. The prefix view does \emph{not} reveal the suffix.
For $\ell=0$, it contributes no supervised positions and the suffix view
reduces to standard anchor-only drafting.
The short-prefix fallback used in our implementation is specified in
Appendix~\ref{app:implementation}.

\paragraph{Why complementarity matters.}
In the basic paired construction, every valid position is supervised once
per sampled block: either under the anchor-only condition or under the
revealed-prefix condition. Without the prefix term, the training signal
would emphasize how to \emph{use} a prefix without directly covering how
to \emph{create} its tokens in that paired example. With uniform weights,
the numerator for one example is exactly the negative log-likelihood of
Equation~\ref{eq:loop-factor}, evaluated with a teacher-forced prefix.
The actual position weights prioritize the start of each stage.
As a result, Equation~\ref{eq:loss} is a weighted surrogate rather than an unweighted
maximum-likelihood objective.



\subsection{Inference and Verification}
\label{sec:inference}

Algorithm~\ref{alg:inference} illustrates one greedy decoding cycle with
$K=1$. For $K>1$, we conduct joint verification: one target forward processes all branches using
the shared context KV cache and branch-specific causal attention.
We commit the longest matching prefix and a target correction or bonus
token, retaining the accepted KV entries. Due to the usage of shared context KV cache and branch-specific causal attention, the cost of joint verification is governable under an appropriate number of branches.  For example,
Table~\ref{tab:latency} reports $25.604$ ms verification
latency at $K=4$, versus $25.363$ ms for single-branch DFlash
(only $0.95\%$ higher). 

\begin{algorithm}[htbp]
\caption{\dloop{} inference with greedy verification ($K=1$, one decoding cycle)}
\label{alg:inference}
\small
\begin{algorithmic}[1]
\Require Context $C$, target $T$, shared drafter $q_\theta$, proposal length $m$
\Ensure Verified token extension $U$
\State $Q^{(0)}\gets q_\theta(\cdot\mid C,z^{(0)})$ \Comment{anchor-only pass}
\State $\widetilde{x}_{1:m}\gets\operatorname{Argmax}(Q^{(0)})$, $c_k\gets Q^{(0)}_k(\widetilde{x}_k)$ \Comment{parallel proposals}
\State $\ell\gets\arg\max_{0\leq j<m}r_j$ \Comment{select one prefix us}
\State $Q^{(\ell)}\gets q_\theta(\cdot\mid C,z^{(\ell)}(\widetilde{x}_{1:\ell}))$ \Comment{prefix-conditioned pass}
\State $y\gets(\widetilde{x}_{1:\ell},\operatorname{Argmax}(Q^{(\ell)}_{\ell+1:m}))$ \Comment{retain prefix, replace suffix}
\State $(a,b)\gets\operatorname{VerifyGreedy}(T,C,y)$ \Comment{accepted length and target token}
\If{$a<m$}
    \State $U\gets(y_{1:a},b)$ \Comment{target correction at rejection}
\Else
    \State $U\gets(y_{1:m},b)$ \Comment{target bonus after full acceptance}
\EndIf
\State \Return $U$ \Comment{truncate at EOS or the output-token limit}
\end{algorithmic}
\end{algorithm}

\section{Experiments}
\label{sec:experiments}
\paragraph{Models and Evaluations.}
We conduct experiments with Qwen3-4B and Qwen3-8B
\citep{yang2025qwen3} as target models. For each benchmark, we report
average acceptance length ($\tau$) and end-to-end decoding speedup over
the vanilla autoregressive baseline. We count the target-generated
correction or bonus token in $\tau$, so it measures the average number of
tokens committed per decoding cycle. The speedup is
\begin{equation}
    \eta=\frac{L_{\mathrm{target}}}{L}
    =\frac{\alpha\,L_{\mathrm{target}}}
    {T_{\mathrm{draft}}+T_{\mathrm{verify}}}.
    \label{eq:eval-speedup}
\end{equation}
Here $L_{\mathrm{target}}$ and $L$ are the average per-token decoding
latencies of the autoregressive baseline and speculative decoding,
respectively. The quantity $\alpha=\tau$ denotes the average tokens
committed per cycle. $T_{\mathrm{draft}}$ and $T_{\mathrm{verify}}$ are
the corresponding average drafting and verification times per cycle.
For \dloop{}, costs include both draft passes, branch preparation,
verification, and state updates.

\paragraph{Datasets.}
We train the drafter on PerfectBlend \citep{xu2024perfectblend} prompts
with responses regenerated by the target model to align the training
continuations with its output distribution. We evaluate on eight
benchmarks covering three domains. Mathematical reasoning includes
GSM8K \citep{cobbe2021gsm8k}, MATH-500 \citep{lightman2023verify}, and
AIME25 \citep{opencompass2025aime}. Code generation includes HumanEval
\citep{chen2021humaneval}, MBPP \citep{austin2021mbpp}, and LiveCodeBench
\citep{jain2024livecodebench}. Conversational tasks include MT-Bench
\citep{zheng2023mtbench} and Alpaca \citep{taori2023alpaca}.

\paragraph{Implementation.}
By following \citep{chen2026dflash}, we use a five-layer DFlash drafter with a block size of
$B=16$. For Qwen3-4B and Qwen3-8B, the drafters are initialized from \citet{chen2026dflash}'s pretrained checkpoints. To be fair,  \dloop{} and all baselines are re-trained with 5 epochs using an initial learning rate of 2e-4, and the evaluation is conducted using the final checkpoint with a batch size of 1.  We set $\beta=0.7$ and the training
loss decay scales to $\tau_s=\gamma=7$.   At inference, both passes share the same drafter weights, and we select $K=4$ candidate prefixes for suffix regeneration. More details can be found in 
Appendix~\ref{app:implementation}.

\paragraph{Baselines.}
We compare \dloop{} with vanilla autoregressive decoding and three
representative speculative decoding methods: EAGLE-3
\citep{li2025eagle3}, DFlash \citep{chen2026dflash}, and DSpark
\citep{cheng2026dspark}. EAGLE-3 uses autoregressive drafting with
target-model features. DFlash generates a block in one parallel diffusion
pass. DSpark adds a sequential head to a parallel diffusion drafter to
introduce token dependencies. These baselines cover autoregressive,
parallel, and semi-autoregressive drafting.
We also report a separate comparison with published D$^2$SD results
\citep{zhang2026d2sd}. D$^2$SD uses DFlash for the initial proposal
and a separately trained variable-prefix drafter for suffix regeneration.
\begin{table}[ht]
\centering
\caption{Average acceptance length ($\tau$) and wall-clock speedup over
autoregressive decoding on Qwen3-4B under greedy ($T=0$) and sampling
($T=1$) decoding. DFlash, DSpark, and \dloop{} results are from our runs.
EAGLE-3 results are quoted from Table~1 of
\citet{chen2026dflash} with tree size 60.
Bold indicates the best result among the methods measured in our runs.}
\label{tab:main}
\fontsize{8}{9.5}\selectfont
\setlength{\tabcolsep}{0.65pt}
\newcommand{\resultmethod}[1]{{\fontsize{6}{7}\selectfont #1}}
\renewcommand{\arraystretch}{1.25}
\setlength{\aboverulesep}{0.4ex}
\setlength{\belowrulesep}{0.4ex}
\begin{tabular}{@{}l
    ccc>{\columncolor{tableblue}}c@{\hspace{2.5pt}}
    ccc>{\columncolor{tableblue}}c@{\hspace{3.5pt}}
    ccc>{\columncolor{tableblue}}c@{\hspace{2.5pt}}
    ccc>{\columncolor{tableblue}[\tabcolsep][0pt]}c@{}}
\toprule
& \multicolumn{8}{c}{\textbf{Greedy} ($T=0$)}
& \multicolumn{8}{c}{\textbf{Sampling} ($T=1$)} \\
\cmidrule(lr){2-9} \cmidrule(lr){10-17}
& \multicolumn{4}{c}{Acc.\ Len.} & \multicolumn{4}{c}{Speedup ($\times$)}
& \multicolumn{4}{c}{Acc.\ Len.} & \multicolumn{4}{c}{Speedup ($\times$)} \\
\cmidrule(lr){2-5} \cmidrule(lr){6-9}
\cmidrule(lr){10-13} \cmidrule(lr){14-17}
\textbf{Dataset}
& \resultmethod{DFlash} & \resultmethod{EAGLE-3} & \resultmethod{DSpark} & \resultmethod{\dloop{}}
& \resultmethod{DFlash} & \resultmethod{EAGLE-3} & \resultmethod{DSpark} & \resultmethod{\dloop{}}
& \resultmethod{DFlash} & \resultmethod{EAGLE-3} & \resultmethod{DSpark} & \resultmethod{\dloop{}}
& \resultmethod{DFlash} & \resultmethod{EAGLE-3} & \resultmethod{DSpark} & \resultmethod{\dloop{}} \\
\midrule
GSM8K & 8.51 & 3.77 & 8.83 & \textbf{10.20} & 6.41 & 2.27 & 6.62 & \textbf{6.77} & 6.39 & 3.68 & 6.83 & \textbf{8.96} & 4.82 & 2.12 & 5.12 & \textbf{5.94} \\
MATH-500 & 8.39 & 3.52 & 8.59 & \textbf{9.84} & 6.32 & 2.10 & 6.44 & \textbf{6.53} & 5.86 & 3.44 & 6.26 & \textbf{7.86} & 4.42 & 1.97 & 4.69 & \textbf{5.21} \\
AIME25 & 7.93 & 3.51 & 7.63 & \textbf{9.09} & 5.98 & 2.13 & 5.72 & \textbf{6.03} & 3.87 & 3.20 & 4.24 & \textbf{5.46} & 2.92 & 1.83 & 3.18 & \textbf{3.62} \\
\midrule
HumanEval & 6.60 & 3.47 & 6.88 & \textbf{7.87} & 4.97 & 2.12 & 5.16 & \textbf{5.22} & 5.05 & 3.39 & 5.33 & \textbf{7.03} & 3.81 & 1.94 & 3.99 & \textbf{4.66} \\
MBPP & 6.14 & 3.38 & 6.44 & \textbf{7.43} & 4.63 & 2.02 & 4.83 & \textbf{4.93} & 4.63 & 3.33 & 4.89 & \textbf{6.75} & 3.49 & 1.92 & 3.67 & \textbf{4.48} \\
LiveCodeBench & 6.26 & 3.22 & 6.40 & \textbf{7.34} & 4.72 & 1.90 & 4.80 & \textbf{4.87} & 3.75 & 3.14 & 4.01 & \textbf{5.16} & 2.83 & 1.82 & 3.01 & \textbf{3.42} \\
\midrule
MT-Bench & 4.64 & 3.49 & 4.72 & \textbf{5.47} & 3.50 & 2.04 & 3.54 & \textbf{3.63} & 3.57 & 3.36 & 3.70 & \textbf{4.82} & 2.69 & 1.91 & 2.77 & \textbf{3.20} \\
Alpaca & 3.33 & --- & 3.35 & \textbf{3.76} & \textbf{2.51} & --- & \textbf{2.51} & 2.49 & 2.54 & --- & 2.72 & \textbf{3.52} & 1.91 & --- & 2.04 & \textbf{2.33} \\
\midrule
\textbf{Average} & 6.48 & 3.48 & 6.61 & \textbf{7.63} & 4.88 & 2.08 & 4.95 & \textbf{5.06} & 4.46 & 3.36 & 4.75 & \textbf{6.20} & 3.36 & 1.93 & 3.56 & \textbf{4.11} \\
\bottomrule
\end{tabular}
\end{table}

\subsection{Main Results}
\label{sec:main-results}
Tables~\ref{tab:main} and~\ref{tab:qwen3-8b} report Qwen3-4B and
Qwen3-8B results under greedy ($T=0$) and sampling ($T=1$) decoding.
We measure DFlash, DSpark, and \dloop{} in our experiments. The EAGLE-3
results are quoted from DFlash \citep{chen2026dflash} as a cross-study
reference.

On Qwen3-4B, \dloop{} achieves the longest acceptance length among the
methods evaluated in our runs on all eight datasets at both temperatures, with
average speedups of $5.06\times$ under greedy decoding and $4.11\times$
under sampling. Its relative improvement in average speedup over DSpark
grows from $2.2\%$ at $T=0$ to $15.4\%$ at $T=1$, showing a larger
benefit under sampling. Greedy Alpaca is the only speedup exception
($2.49\times$ versus DSpark's $2.51\times$), where longer acceptance
does not fully offset the additional drafting cost.

On Qwen3-8B, \dloop{} improves both acceptance length and speedup over
DFlash and DSpark on every dataset at both temperatures. It achieves
average acceptance lengths of $7.19$ and $5.77$, with average speedups
of $5.02\times$ and $4.03\times$ under greedy and sampling decoding,
respectively. The relative gain in average speedup over DSpark increases
from $9.7\%$ at $T=0$ to $26.2\%$ at $T=1$, consistent with the larger
sampling gains observed on Qwen3-4B. Longer accepted drafts thus offset
the additional pass across math, code, and chat tasks.

\begin{table}[!t]
\centering
\caption{Average acceptance length ($\tau$) and wall-clock speedup over
autoregressive decoding on Qwen3-8B under greedy ($T=0$) and sampling
($T=1$) decoding. DFlash, DSpark, and \dloop{} results are from our runs.
EAGLE-3 results are quoted from Table~1 of
\citet{chen2026dflash} with tree size 60.
Bold indicates the best result among the methods measured in our runs.}
\label{tab:qwen3-8b}
\fontsize{8}{9.5}\selectfont
\setlength{\tabcolsep}{0.65pt}
\newcommand{\resultmethod}[1]{{\fontsize{6}{7}\selectfont #1}}
\renewcommand{\arraystretch}{1.25}
\setlength{\aboverulesep}{0.4ex}
\setlength{\belowrulesep}{0.4ex}
\begin{tabular}{@{}l
    ccc>{\columncolor{tableblue}}c@{\hspace{2.5pt}}
    ccc>{\columncolor{tableblue}}c@{\hspace{3.5pt}}
    ccc>{\columncolor{tableblue}}c@{\hspace{2.5pt}}
    ccc>{\columncolor{tableblue}[\tabcolsep][0pt]}c@{}}
\toprule
& \multicolumn{8}{c}{\textbf{Greedy} ($T=0$)}
& \multicolumn{8}{c}{\textbf{Sampling} ($T=1$)} \\
\cmidrule(lr){2-9} \cmidrule(lr){10-17}
& \multicolumn{4}{c}{Acc.\ Len.} & \multicolumn{4}{c}{Speedup ($\times$)}
& \multicolumn{4}{c}{Acc.\ Len.} & \multicolumn{4}{c}{Speedup ($\times$)} \\
\cmidrule(lr){2-5} \cmidrule(lr){6-9}
\cmidrule(lr){10-13} \cmidrule(lr){14-17}
\textbf{Dataset}
& \resultmethod{DFlash} & \resultmethod{EAGLE-3} & \resultmethod{DSpark} & \resultmethod{\dloop{}}
& \resultmethod{DFlash} & \resultmethod{EAGLE-3} & \resultmethod{DSpark} & \resultmethod{\dloop{}}
& \resultmethod{DFlash} & \resultmethod{EAGLE-3} & \resultmethod{DSpark} & \resultmethod{\dloop{}}
& \resultmethod{DFlash} & \resultmethod{EAGLE-3} & \resultmethod{DSpark} & \resultmethod{\dloop{}} \\
\midrule
GSM8K & 7.35 & 3.71 & 7.42 & \textbf{9.46} & 5.58 & 2.23 & 5.63 & \textbf{6.61} & 5.28 & 3.59 & 5.74 & \textbf{8.24} & 4.01 & 2.07 & 4.36 & \textbf{5.76} \\
MATH-500 & 7.62 & 3.49 & 8.04 & \textbf{9.97} & 5.79 & 2.05 & 6.10 & \textbf{6.96} & 5.21 & 3.38 & 5.71 & \textbf{7.79} & 3.96 & 1.94 & 4.34 & \textbf{5.44} \\
AIME25 & 7.07 & 3.44 & 7.32 & \textbf{8.69} & 5.37 & 2.05 & 5.56 & \textbf{6.07} & 3.62 & 3.18 & 3.93 & \textbf{5.41} & 2.75 & 1.84 & 2.98 & \textbf{3.78} \\
\midrule
HumanEval & 5.99 & 3.65 & 6.36 & \textbf{7.41} & 4.55 & 2.17 & 4.83 & \textbf{5.18} & 4.14 & 3.54 & 4.59 & \textbf{6.36} & 3.15 & 2.05 & 3.48 & \textbf{4.44} \\
MBPP & 5.19 & 3.25 & 5.77 & \textbf{6.39} & 3.94 & 1.93 & 4.38 & \textbf{4.46} & 3.87 & 3.16 & 4.28 & \textbf{5.67} & 2.94 & 1.85 & 3.25 & \textbf{3.96} \\
LiveCodeBench & 5.39 & 3.03 & 5.88 & \textbf{6.75} & 4.10 & 1.81 & 4.46 & \textbf{4.71} & 3.26 & 2.92 & 3.55 & \textbf{4.73} & 2.48 & 1.72 & 2.70 & \textbf{3.30} \\
\midrule
MT-Bench & 4.34 & 3.26 & 4.31 & \textbf{5.20} & 3.30 & 1.90 & 3.27 & \textbf{3.63} & 3.10 & 3.05 & 3.31 & \textbf{4.53} & 2.36 & 1.70 & 2.51 & \textbf{3.16} \\
Alpaca & 3.21 & --- & 3.16 & \textbf{3.67} & 2.44 & --- & 2.40 & \textbf{2.56} & 2.49 & --- & 2.50 & \textbf{3.39} & 1.89 & --- & 1.90 & \textbf{2.37} \\
\midrule
\textbf{Average} & 5.77 & 3.40 & 6.03 & \textbf{7.19} & 4.38 & 2.02 & 4.58 & \textbf{5.02} & 3.87 & 3.26 & 4.20 & \textbf{5.77} & 2.94 & 1.88 & 3.19 & \textbf{4.03} \\
\bottomrule
\end{tabular}
\end{table}

\subsection{Analysis of Drafting Efficiency}
\label{sec:efficiency-analysis}
\begingroup
\setlength{\intextsep}{5pt}
\begin{wrapfigure}{r}{0.64\textwidth}
\centering
\begin{minipage}[b]{0.515\linewidth}
    \centering
    \includegraphics[width=\linewidth]{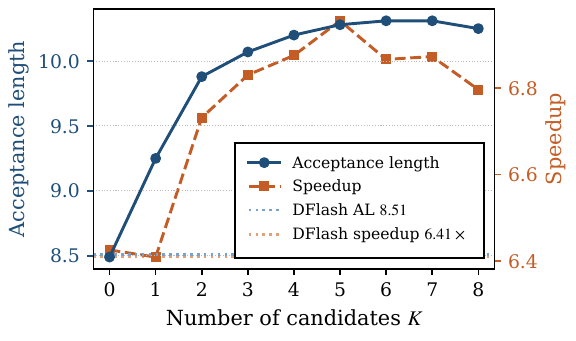}
    \par\vspace{2pt}
    {\footnotesize (a) Candidate budget.\par}
\end{minipage}\hfill
\begin{minipage}[b]{0.475\linewidth}
    \centering
    \includegraphics[width=\linewidth]{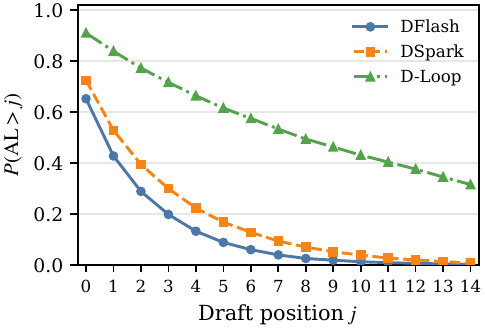}
    \par\vspace{2pt}
    {\footnotesize (b) Acceptance by position.\par}
\end{minipage}
\setlength{\abovecaptionskip}{3pt}
\caption{Qwen3-4B on GSM8K: (a) acceptance length and speedup versus $K$;
(b) acceptance prob. over position $j$.}
\label{fig:ablation-latency}
\end{wrapfigure}

Figure~\ref{fig:ablation-latency}(a,b) examines drafting efficiency on
GSM8K using Qwen3-4B.
(a) \textbf{A small candidate budget already improves performance.}
Increasing $K$ beyond the single-pass setting ($K=0$) improves acceptance
until it saturates, with speedup peaking at $K=5$. Gains in both metrics
are already evident at $K=2$ and $K=3$;
(b) \textbf{\dloop{} retains more accepted tokens toward the suffix.}
It maintains a higher and more slowly decaying $P(\mathrm{AL}>j)$ than
DFlash and DSpark. The separation persists toward the end of the block,
indicating longer useful prefixes. This cumulative survival measure
reflects both early rejections and later-token quality, rather than
isolated per-position accuracy.
\par\endgroup

Table~\ref{tab:dloop-d2sd} compares \dloop{} with published D$^2$SD
results and ablates complementary prefix--suffix training on Qwen3-8B. Some obvervations can be found as follows:
(a) \textbf{A single drafter with a looped structure is sufficient.}
\dloop{} uses one five-layer drafter, compared with two in D$^2$SD (which equals to 10-layer drafter). However, \dloop{} can beat D$^2$SD in most cases, leading to higher average acceptance and speedup performance. (b) \textbf{The
complementary prefix–suffix objective is effective.}
With the same architecture, suffix-only training supervises
prefix-conditioned prediction, while the full loss also trains anchor-only
prefix prediction for the initial proposal. The full loss raises mean acceptance
from $7.95$ to $8.11$ at $T=0$ and from $6.26$ to $6.37$ at $T=1$.
Mean speedups improve from $5.60\times$ to $5.67\times$ and from
$4.40\times$ to $4.45\times$, respectively. Acceptance improves on five
of six datasets at $T=0$ and four at $T=1$. 
\begin{table}[ht]
\centering
\caption{Qwen3-8B comparison and loss ablation (acceptance length / speedup
in $\times$). Bold marks metric maxima. D-Loop-suffix-loss uses suffix
supervision only. \dloop{} uses the full complementary prefix--suffix loss.
D$^2$SD results are from \citet{zhang2026d2sd}, Table~3 ($\gamma=16$, $K=4$),
using two five-layer drafters ($5+5=10$). Our two variants each share one
five-layer drafter.}
\label{tab:dloop-d2sd}
\fontsize{8}{9.5}\selectfont
\setlength{\tabcolsep}{1pt}
\renewcommand{\arraystretch}{1.2}
\setlength{\aboverulesep}{0.4ex}
\setlength{\belowrulesep}{0.4ex}
\begin{tabular}{clc|c|c|c|c|c|c|c}
\toprule
\textbf{Decoding} & \textbf{Method} & \shortstack{\textbf{\# Draft}\\\textbf{Layers}}
& GSM8K & MATH & AIME25 & HumanEval & MBPP & LCB & \textbf{Average} \\
\midrule
& D$^2$SD & 10 & 9.21 / 6.54 & 8.56 / 6.03 & 7.26 / 5.10
& \textbf{7.47} / \textbf{5.37} & \textbf{6.96} / \textbf{5.03}
& 6.44 / 4.62 & 7.65 / 5.45 \\
& D-Loop-suffix-loss & 5
& 9.12 / 6.43 & 9.62 / 6.78 & 8.57 / 6.04 & 7.34 / 5.18
& 6.36 / 4.49 & 6.68 / 4.67 &  7.95 / 5.60 \\
\rowcolor{tableblue}
\cellcolor{white}\multirow{-3}{*}{\shortstack{Greedy\\($T=0$)}}
& \dloop{}-full-loss & 5 & \textbf{9.46} / \textbf{6.61} & \textbf{9.97} / \textbf{6.96}
& \textbf{8.69} / \textbf{6.07} & 7.41 / 5.18 & 6.39 / 4.46
& \textbf{6.75} / \textbf{4.71} & \textbf{8.11} / \textbf{5.67} \\
\midrule
& D$^2$SD & 10 & 7.38 / 5.32 & 6.81 / 4.90
& \textbf{5.75} / \textbf{4.11} & 5.50 / 4.05 & 4.88 / 3.24
& \textbf{4.90} / \textbf{3.53} & 5.87 / 4.19 \\
& D-Loop-suffix-loss & 5
& 8.23 / \textbf{5.80} & 7.60 / 5.30 & 5.48 / 3.87
& 6.07 / 4.28 & 5.59 / 3.94 & 4.58 / 3.23 & 6.26 / 4.40 \\
\rowcolor{tableblue}
\cellcolor{white}\multirow{-3}{*}{\shortstack{Sampling\\($T=1$)}}
& \dloop{}-full-loss & 5 & \textbf{8.24} / 5.76 & \textbf{7.79} / \textbf{5.44}
& 5.41 / 3.78 & \textbf{6.36} / \textbf{4.44} & \textbf{5.67} / \textbf{3.96}
& 4.73 / 3.30 & \textbf{6.37} / \textbf{4.45} \\
\bottomrule
\end{tabular}
\vspace{-0.4cm}
\end{table}

\begingroup
\setlength{\intextsep}{4pt}
\setlength{\columnsep}{10pt}
\begin{wraptable}{r}{0.48\textwidth}
\centering
\setlength{\abovecaptionskip}{0pt}
\caption{Latency on GSM8K with Qwen3-8B. $t_{\mathrm{ar}}$, $t_{\mathrm{draft}}$, and $t_{\mathrm{verify}}$ mean the time consumption for autoregressive token-by-token decoding, drafting, and target verification. }
\label{tab:latency}
\fontsize{8}{9.5}\selectfont
\setlength{\tabcolsep}{2pt}
\renewcommand{\arraystretch}{1.3}
 \begin{tabular}{lccccc}
        \toprule
        Method & $t_{\mathrm{ar}}$ & $t_{\mathrm{draft}}$ & $t_{\mathrm{verify}}$
        & AL & Speedup \\
        & ms/tok & ms/iter & ms/iter & & \\
        \midrule
        DFlash & 22.29 & 4.208 & 25.363 & 5.28 & $3.98\times$ \\
        DSpark & 22.29 & 4.208 & 25.531 & 5.74 & $4.30\times$ \\
        \rowcolor{tableblue}
        \dloop{} & 22.29 & 9.428 & 25.604 & \textbf{8.24}
        & $\mathbf{5.24}\times$ \\
        \bottomrule
    \end{tabular}
\end{wraptable}

Table~\ref{tab:latency} shows that
\textbf{longer acceptance offsets the additional drafting cost.}
At $K=4$, drafting time rises from $4.208$ to $9.428$ ms per iteration,
while verification stays near $25.5$ ms. Longer acceptance spreads this
verification cost over more committed tokens, yielding higher end-to-end
speedup than DFlash and DSpark. Thus, the extra forward pass improves
speedup when its acceptance gain offsets the additional drafting latency.
\par\endgroup

\FloatBarrier

\vspace{-0.3cm}
\section{Related Work}
\label{sec:related-work}
\vspace{-0.2cm}
\textbf{Speculative drafting and refinement.}
Speculative decoding verifies inexpensive proposals with the target
\citep{leviathan2023speculative,chen2023sampling}.
EAGLE-3 \citep{li2025eagle3} uses target features for autoregressive
drafting, while DFlash \citep{chen2026dflash} predicts blocks in parallel.
DSpark \citep{cheng2026dspark} and Domino \citep{huang2026domino}
add lightweight causal modules that condition corrections on preceding
tokens. xPress \citep{wang2026xpress} performs this refinement through
parallel Jacobi updates. These methods assign causal refinement to an
additional module.
D$^2$SD \citep{zhang2026d2sd} selects candidate prefixes using token
confidence and regenerates their suffixes with a separate variable-prefix
drafter. It also evaluates direct DFlash reuse.
\dloop{} trains one shared backbone for both initial proposal and
redrafting through complementary prefix--suffix supervision, exposing
the proposed prefix to the full drafter during refinement.

\textbf{Dependencies in parallel generation.}
Non-autoregressive translation studies repetition and multimodality
errors \citep{gu2018nonautoregressive,ran2020recover}.
We characterize repetition in speculative drafts, where early rejection
also discards the suffix. Block diffusion \citep{arriola2025block}
conditions on preceding blocks while retaining parallel prediction
within each block. \dloop{} applies this principle within a speculative
block. Recurrent-depth models \citep{geiping2025recurrent} share parameters
across latent computation steps. Our loop conditions on discrete proposal
tokens, adding computation without a second set of drafter weights.

\vspace{-0.5cm}
\section{Conclusion}
\label{sec:conclusion}
We identify and analyze a repetition trap in marginal diffusion drafting.
\dloop{} introduces intra-block causal conditioning by reusing one
drafter for initial proposal and prefix-conditioned suffix generation,
supported by complementary prefix--suffix training.
Experiments on Qwen3-4B and Qwen3-8B show longer accepted drafts and
improved decoding speed across math, code, and chat tasks.
These results support improving draft dependencies without a separate
refinement model.

\textbf{Limitations and future work.} Our fixed looped schedule may be inefficient when acceptance is much
short. On greedy Alpaca with Qwen3-4B, the acceptance gain does not offset the extra pass, slightly
reducing speedup. Future work could adapt the number of passes to expected acceptance gains and
latency, skipping refinement when its return is low and adding passes when worthwhile.

\bibliographystyle{iclr2027_conference}
\bibliography{references}

\appendix
\section{Derivations and Implementation Clarifications}
\label{app:proofs}

%

\subsection{Why Intra-Block Causal Conditioning Helps}
\label{sec:information}

In Equation~\ref{eq:parallel}, every position predicts from the same
verified context without observing the tokens selected at earlier
positions. Each marginal therefore averages over possible continuations,
even after the first pass has produced a particular prefix.
\dloop{} exposes that prefix to the backbone, allowing suffix predictions
to adapt to the continuation already proposed. Bidirectional attention
over masked positions alone cannot provide this realized-token information.

To isolate this modeling advantage, fix a split $\ell$ independently of
the proposal tokens. Let $q^\star_{\mathrm{one}}$ be
Equation~\ref{eq:parallel} with exact target marginals, and let
$q^\star_{\mathrm{loop}}$ be Equation~\ref{eq:loop-factor} with the same
prefix marginals and exact prefix-conditioned suffix probabilities.
Their expected log-likelihood difference, evaluated on target
continuations and their contexts, is
\begin{equation}
    \E_p\!\left[\log
    \frac{q^\star_{\mathrm{loop}}(X_{1:m}\mid C)}
         {q^\star_{\mathrm{one}}(X_{1:m}\mid C)}\right]
    =\sum_{k=\ell+1}^{m}I(X_k;X_{1:\ell}\mid C)\geq0.
    \label{eq:causal-advantage}
\end{equation}
Each term measures information that the prefix provides about
one suffix token beyond the verified context. Equivalently, it is the
reduction in that token's uncertainty from $H(X_k\mid C)$ to
$H(X_k\mid X_{1:\ell},C)$. The prefix factors are identical and cancel.
Therefore, the gain comes entirely from replacing context-only suffix marginals
with prefix-conditioned predictions (Appendix~\ref{app:prefix}). It is
strictly positive whenever at least one suffix token depends on the
prefix given $C$.

Thus, intra-block causal conditioning can recover dependencies that
Equation~\ref{eq:parallel} discards, while keeping suffix prediction
parallel. This comparison describes the advantage with exact predictions.
In practice, a trained shared drafter must learn both prediction conditions to realize
it. Dependencies within each parallel stage remain unresolved, and a
likelihood gain alone does not guarantee longer acceptance. The
complementary objective in Section~\ref{sec:training} trains the same
backbone for both passes.

\subsection{Likelihood Advantage of Prefix Conditioning}
\label{app:prefix}
For a fixed split $\ell$, define the ideal distributions by
\begin{equation}
\begin{aligned}
    q^\star_{\mathrm{one}}(x_{1:m}\mid C)
    &=\prod_{k=1}^{m}p(x_k\mid C),\\
    q^\star_{\mathrm{loop}}(x_{1:m}\mid C)
    &=\prod_{k=1}^{\ell}p(x_k\mid C)
      \prod_{k=\ell+1}^{m}p(x_k\mid x_{1:\ell},C).
\end{aligned}
\end{equation}
These minimize expected negative log-likelihood within the respective
factorizations when each component distribution is unrestricted.
The common prefix factors cancel in their ratio. Taking expectations
under the target joint distribution gives
\begin{equation}
\begin{aligned}
    \E_p\!\left[\log
    \frac{q^\star_{\mathrm{loop}}(X_{1:m}\mid C)}
         {q^\star_{\mathrm{one}}(X_{1:m}\mid C)}\right]
    &=\sum_{k=\ell+1}^{m}\E_p\!\left[\log
      \frac{p(X_k\mid X_{1:\ell},C)}{p(X_k\mid C)}\right]\\
    &=\sum_{k=\ell+1}^{m}I(X_k;X_{1:\ell}\mid C).
\end{aligned}
\end{equation}
Nonnegativity follows from that of conditional mutual information. The
same quantity equals
$\E_C[\KL(p\|q^\star_{\mathrm{one}})-\KL(p\|q^\star_{\mathrm{loop}})]$,
because the target entropy cancels. The identity concerns likelihood
under the target distribution, not the distribution of generated
prefixes or token-dependent split selection. It does not assert that a
finite shared network attains both ideal distributions, nor that a
likelihood improvement necessarily improves greedy acceptance.

\subsection{Training Implementation and Short-Prefix Fallback}
\label{app:implementation}
\paragraph{Complementary views.}
For a sampled prefix length $\ell_i\geq2$, both training views use the
same anchor $a_i$. The prefix view reveals only the anchor and supervises
positions $1,\ldots,\ell_i$. The suffix view additionally reveals the
ground-truth prefix and supervises positions $\ell_i+1,\ldots,m$.
This is the paired construction in Section~\ref{sec:training}.

\paragraph{Fallback for $\ell_i<2$.}
An anchor-only prefix view would receive no supervised token when
$\ell_i=0$ and only one when $\ell_i=1$. To make better use of this
forward pass, we replace that view with a full anchor-only training
block at a neighboring sampled anchor in the same sequence. The suffix
view remains at $a_i$ with its original sampled prefix length.
The neighboring anchor is obtained by cyclically shifting the sampled
anchor array by one position, as in
\texttt{torch.roll(anchor\_positions, 1, dims=1)}.
For anchors $(a_1,\ldots,a_N)$, this gives $(a_N,a_1,\ldots,a_{N-1})$.
Thus, ``neighboring'' refers to the previous entry in the sampled anchor
array, with wraparound, rather than necessarily the preceding token.

Let $\pi(i)$ denote this cyclic predecessor. The selected prefix-view
anchor and its loss weights, for $k=1,\ldots,m$, are
\begin{equation}
    a_i^P=\begin{cases}
       a_i,&\ell_i\geq2,\\
       a_{\pi(i)},&\ell_i<2,
    \end{cases}
    \qquad
    \widetilde w^P_{ik}
    =v^P_{ik}e^{-(k-1)/\gamma}
    \begin{cases}
       \Ind\{k\leq\ell_i\},&\ell_i\geq2,\\
       1,&\ell_i<2.
    \end{cases}
    \label{eq:short-prefix-fallback}
\end{equation}
Here $v^P_{ik}$ excludes invalid or padded prediction positions, and
$\gamma$ is the anchor-relative loss decay scale. The context and
ground-truth labels are taken from the selected anchor $a_i^P$.
Its input contains that anchor followed by $m$ mask tokens. In the
fallback case, all valid prediction positions receive supervision with
the usual DFlash decay. For example, with $B=16$ and $\ell_i=1$,
the suffix view supervises positions $2,\ldots,15$ at the original
anchor. The replacement prefix view supervises all 15 proposal positions
at the neighboring anchor. The anchor token itself is never supervised.

\paragraph{Loss accumulation and view isolation.}
The implementation processes the two views in separate forward passes
with shared drafter parameters. Each view is divided into anchor chunks
with gradient checkpointing to reduce peak activation memory. Weighted
cross-entropy sums and weight sums are accumulated across both views
before applying the normalization in Equation~\ref{eq:loss}, using
$\epsilon=10^{-6}$. For a fallback block, the prefix terms use
$a_i^P$ and $\widetilde w^P_{ik}$ from
Equation~\ref{eq:short-prefix-fallback}.
The anchor-only view cannot access the revealed prefix of its companion
view. Target features are restricted to the context available at the
anchor of each view.

In the basic paired construction, each prediction position is supervised
in exactly one view before weighting. The fallback instead mixes
prefix-conditioned suffix training with full anchor-only training on
another sampled block. Therefore, this exact partition no longer holds
for short-prefix blocks. This training fallback adds no model parameters
or inference passes.

\subsection{Correctness of Greedy and Stochastic Verification}
\label{app:verification}
\paragraph{Deterministic drafting at both temperatures.}
Both draft passes use position-wise argmax at $T=0$ and $T=1$.
Prefix selection also remains deterministic. We compute first-pass
token confidences from the unscaled draft softmax and select the
top-$K$ prefix lengths using the score in Section~\ref{sec:loop}.
Here top-$K$ ranks prefix lengths, not vocabulary tokens.
For the same context and model state, the initial proposal, selected
prefixes, and regenerated suffixes are therefore identical at both
temperatures. During a drafting--verification cycle, temperature affects
only the target's token selection.

\paragraph{Greedy decoding.}
Let $g(C)$ be the target's next-token argmax with a deterministic
tie-breaking rule. A branch $y_{1:m}$ is verified by comparing $y_k$ to
$g(C,y_{<k})$. Before its first disagreement, induction gives that every
$y_k$ equals the corresponding token of the target's greedy continuation.
Therefore all branches' accepted prefixes are prefixes of the same
target continuation, and choosing the longest one cannot change that
continuation. At its next position, a target correction or bonus token
also equals the next target token. Repeating the argument cycle by cycle
proves equality of the complete greedy output, subject to correct
branch-specific causal attention, cache rollback, and numerical
evaluation of the target.

\paragraph{Target token selection.}
Let $a_v(h)$ be the target logit for token $v$ given output history $h$,
and let $g(h)$ be its argmax with fixed tie-breaking. The implementation
selects a token according to
\begin{equation}
    p_T(v\mid h)=
    \begin{cases}
        \Ind\{v=g(h)\}, & T<10^{-5},\\[2pt]
        \displaystyle\frac{\exp(a_v(h)/T)}
        {\sum_{u\in\mathcal V}\exp(a_u(h)/T)}, & T\geq10^{-5},
    \end{cases}
    \label{eq:target-sampling-rule}
\end{equation}
where $\mathcal V$ is the vocabulary. The first case is implemented by
argmax. Otherwise, we draw one categorical sample from the temperature-scaled
softmax, without top-$k$ or nucleus (top-$p$) filtering. The initial
target-generated token and correction or bonus tokens use this same rule.

\paragraph{One shared draw across surviving branches.}
 Joint verification computes target logits for
each branch under its own causal history. At position $k$, the surviving
branches all match the previously sampled tokens $y_{<k}$ and thus share
the target distribution $p_T(\cdot\mid C,y_{<k})$. We use the logits
of the first surviving branch to draw a single token $y_k$. Every
surviving branch is compared against this same draw, and those with a
different proposal at position $k$ are discarded. If none remains,
the already sampled $y_k$ becomes the correction token without
resampling. The next cycle starts from the resulting output history.
If a complete draft survives, the target draws a bonus token using
Equation~\ref{eq:target-sampling-rule}.

\paragraph{Preservation of the target distribution.}
At each step, the emitted token follows $p_T(\cdot\mid h)$ for the
actual output history $h$. By induction, the generated sequence has
the same distribution as autoregressive target sampling at the same
temperature without vocabulary filtering. Deterministic proposal and
prefix selection affect how far cached verification can continue, but
do not change this conditional distribution. This argument requires
causal branch isolation, consistent position indices, correct KV-cache
updates, and one shared target draw per position across surviving branches.

\paragraph{Effect on acceptance length.}
For a fixed history $h$ and proposal token $v$, stochastic matching
occurs with probability $p_T(v\mid h)$. With multiple surviving branches,
the probability of continuing verification is the total target probability
mass of their distinct next-token proposals. A token accepted under
greedy verification can therefore be rejected under sampling, while a
non-argmax proposal can be accepted. Once a sampled token differs from
the greedy continuation, subsequent cycles also receive different
contexts and may produce different proposals. The observed reduction in
acceptance length at $T=1$ thus reflects stochastic target matching and
diverging generation histories, rather than randomness in the drafter.
It is not a universal ordering of acceptance lengths across temperatures.
\end{document}